\documentclass{article}
\usepackage[font=small]{caption}
\usepackage[labelformat=empty, position=top]{subcaption}
\usepackage{graphicx}
\usepackage[export]{adjustbox}
\usepackage[utf8]{inputenc}
\usepackage{array}
\usepackage{diagbox}
\usepackage{booktabs}
\usepackage{mathrsfs}
\usepackage{makecell}
\usepackage{tikz}
\usepackage{amsmath}
\usepackage{amsthm}
\usepackage{amssymb}
\usepackage[rightcaption]{sidecap}
\usepackage[nonatbib, preprint]{neurips_2023}
\usepackage[backend=biber,style=numeric-comp,sortcites,sorting=none,backref=false]{biblatex}
\usepackage{hyperref}
\renewbibmacro{in:}{}

\title{Physics Inspired Approaches To Understanding Gaussian Processes}
\author{
  Maximilian P.~Niroomand \\
  University of Cambridge\\
  Cambridge, UK \\
  \texttt{mpn26@cam.ac.uk} \\
  \And
  Luke Dicks \\
  IBM Research Europe \\
  Daresbury, UK \\
  \And
  Edward O.~Pyzer-Knapp \\
  IBM Research Europe \\
  Daresbury, UK \\
  \And
  David J.~Wales \\
  University of Cambridge\\
  Cambridge, UK \\
}

\begin{document}

\maketitle

\section{Abstract}
Prior beliefs about the latent function to shape inductive biases can be incorporated into a Gaussian Process (GP) via the kernel. However, beyond kernel choices, the decision-making process of GP models remains poorly understood. In this work, we contribute an analysis of the loss landscape for GP models using methods from physics. We demonstrate $\nu$-continuity for Mat\'ern kernels and outline aspects of catastrophe theory at critical points in the loss landscape. By directly including $\nu$ in the hyperparameter optimisation for Mat\'ern kernels, we find that typical values of $\nu$ are far from optimal in terms of performance, yet prevail in the literature due to the increased computational speed. We also provide an \textit{a priori} method for evaluating the effect of GP ensembles and discuss various voting approaches based on physical properties of the loss landscape. The utility of these approaches is demonstrated for various synthetic and real datasets. Our findings provide an enhanced understanding of the decision-making process behind GPs and offer practical guidance for improving their performance and interpretability in a range of applications.

\iffalse
Prior beliefs about the latent function to shape inductive biases can be incorporated into a Gaussian Process (GP) via the choice of kernel. Choosing an appropriate kernel is perhaps the most significant task in Gaussian Process regression. Yet, the kernel decision making process for GP machine learning is surprisingly poorly understood. Amongst the standard repertoire of kernels, the Mat\'{e}rn kernel is the most versatile. Specified by a hyperparameter $\nu$ that controls the smoothness, it can take the form of various other kernels such as the popular RBF or exponential kernel. By explicitly incorporating $\nu$ into hyperparameter optimisation, we find that typical values of $\nu$ are far from optimal in terms of performance, yet prevail in the literature due to the increased computational speed. We also show that in $\nu$-adjacent loss landscapes, corresponding minima for subsequent values of $\nu$ transition smoothly. Lastly, we show how information about the loss landscape can be incorporated in GP ensembles and 
\fi

\section{Introduction}
The interpretation and understanding of model decision-making is one of the most important unsolved problems in machine learning. Practitioners in fields from healthcare to cybersecurity remain reluctant to deploy machine learning solutions whilst the decision process remains opaque. Gaussian Processes (GPs) are a well-known class of nonparametric, supervised learning models that introduce a Bayesian framework to machine learning. The Bayesian framework allows the construction of a confidence measure around predictions, and therefore GPs overcome one of the most prominent limitations of other, non-Bayesian methods, namely uncertainty quantification. While uncertainty quantification provides an important step towards model reliability, the more fundamental issue of interpretability remains unsolved, and a significant barrier to the widespread adoption of GPs in critical and sensitive applications.
%Because of the encouraging advantages over non-Bayesian methods, solving interpretability is of critical important towards the widespread adoption of GPs in critical and sensitive real-life applications.  

Mathematically, a GP is a collection of random variables, any finite subset of which is jointly Gaussian distributed \cite{williams2006gaussian}. Unlike a neural network, where the learned parameters predict some output in combination with the functional form of the model, a GP learns a distribution over functions to fit the data. In parametric machine learning, a distribution $p(y|\mathcal{X},\theta)$ is employed to explain some input data $\mathcal{X}$, with the optimal set of coefficients $\theta$ being inferred by maximum likelihood or maximum \textit{a posteriori} estimation. In nonparametric methods, such as Gaussian processes, a weighted distribution over functions $p(f|\mathcal{X}) = \mathcal{N}(f|\mu, \sigma)$ is learned, where the parameters are the function itself. Given some data, a posterior distribution $p(f_\ast|\mathcal{X}_\ast,\mathcal{X},f)$ for new predictions $f_\ast$ at position $\mathcal{X}_\ast$ can be computed. 

Gaussian processes allow the incorporation of inductive biases via the covariance kernel. The choice of kernel, together with a mean function $\mu(x)$ (which is usually just set to 0 such that $\mu(x) = 0$), is the critical design choice when constructing a GP. However, the number of kernels being used in practice is surprisingly small. Driven partly by their availability in standard GP Python packages, but more so by convention, there are perhaps a handful of different kernels that are commonly used. We want to explore alternative kernels and provide insights into the benefits and disadvantages of using non-standard choices. In particular, we study the Mat\'{e}rn kernel, a general formulation that encompasses most other popular kernels by variation of its smoothness parameter $\nu$.

Obtaining an uncertainty estimate in GPs does not come for free and the most prominent disadvantage of using GPs over parametric machine learning methods is computational cost. The training of GPs scales naively as $\mathcal{O}(n^3)$ for $n$ data points due to inversion of the covariance matrix. Various approaches exist to reduce the complexity \cite{liu2020gaussian}, which usually revolve around the concept of sparse GPs \cite{lawrence2002fast} and inducing point methods \cite{titsias2009variational}. However, with computational resources continuing to grow, and an increased demand for explainable machine learning, the growing interest in GPs will likely continue. 

The study of loss landscapes in machine learning aims to increase interpretability of black box models and identify new ways to think about machine learning. Due to the computational cost associated with evaluating large areas of the loss landscape, relatively little work exists on characterisation \cite{sun2020global} and visualisation \cite{bosman2020visualising, li2018visualizing}. In this work, we study loss landscapes using tools from the physical sciences \cite{wales2003energy}, where molecular energy landscapes are employed to predict thermodynamic and kinetic properties of a system. We will treat ML loss landscapes analogously, as in previous work \cite{verpoort2020archetypal, niroomand2022capacity, dicks2022elucidating}, which identified various similarities between molecular energy landscapes and machine learning loss landscapes.

In GPs, loss landscapes are a graph encoding of the loss function, usually the negative log marginal likelihood (-lml). Since GPs are a non-parametric method, the training phase does not include learning a set of model parameters as for neural networks. Rather, the solution space of the loss function and hence its arguments, is the set of hyperparameters $\theta \in \Theta$ that characterise the kernel. %The smoothness parameter $\nu$ in Matern kernels is an example of such a hyperparameter, yet it is usually fixed before training for computational reasons which are explained below. However, this restriction is unnecessary and we will additionally examine the effect of including $\nu$ in the hyperparameter space $\Theta$. 
Current research in loss landscapes focuses almost exclusively on supervised parametric methods. Such landscapes usually exhibit rich and high-dimensional structures due to the vast number of parameters. Loss landscapes of non-parametric methods such as GPs are inherently simpler, due to a substantially lower number of parameters. In a GP loss landscape, the global minimum represents the set of kernel hyperparameters $\theta$ that provide the most accurate solution given training data. Understanding the solution space for a GP therefore provides an understanding of robustness and inference accuracy of the underlying GP model.  

\subsection{Related work}
\textbf{Loss landscapes:}
The study of loss landscapes in machine learning has received increasing attention in recent years. Insights into the structure and geometric properties of the loss landscape has been employed to increase model performance \cite{niroomand2022capacity, fort2019deep}, adversarial robustness \cite{eustratiadis2022attacking}, interpretability \cite{niroomand2022characterising, niroomand2023physics}, and generalisability \cite{baldassi2021unveiling, ruiz2021tilting, choromanska2015loss} of neural networks. Empirically, loss landscape characteristics have also been considered to explain fundamental aspects of machine learning, such as the quality of minima \cite{wu2017towards} and the structure of the solution space as a function of hyperparameters \cite{verpoort2020archetypal}. Unlike parametric methods, the parameters in GPs are the function itself \cite{williams2006gaussian} and only a selected few  hyperparameters specifying the kernel must be learned. Previous work has considered loss landscapes in the context of Bayesian optimisation. However, we believe that the present contribution is the first account of a detailed study of GP loss landscapes.
\newline \textbf{GP interpretability:} The interpretation of machine learning models is one of the most active fields of research today \cite{rudin2022interpretable}. In GPs specifically, various approaches have been studied to interpret inference beyond kernel choices. Previous approaches include kernel decomposition methods and local explanations to mirror linear models \cite{cheng2019additive, yoshikawa2021gaussian}, while others employ sensitivity analysis, a more classical approach used across various machine learning methods \cite{martinez2020crop}. Yet, to the best of our knowledge, no methods exist that consider the loss landscape or use well-established methods from physics towards GP interpretability.
\newline\textbf{GP ensemble methods:} Ensemble methods that combine multiple predictors by methods such as bagging \cite{breiman1996bagging}, are commonly used to improve model performance. The intuition that a combination of single predictors should outperform a single model has been confirmed theoretically and empirically \cite{lakshminarayanan2017simple, niroomand2022capacity}. Both papers ascribe the performance increase in bagging ensembles to a set of qualitatively different modes of the loss landscape, akin to different minima. GP ensembles have been used in online learning applications \cite{lu2020ensemble}, usually with Bayesian ensemble weights \cite{deng2022deep}, and as GP experts with different kernel functions \cite{lu2022incremental}. However, widespread adoption of GP ensembles is impeded by high computational costs and often only minor improvements in terms of inference accuracy. In contrast to neural network ensembles, none of the above methods consider features of the loss landscapes, as, for example, suggested in \textcite{fort2019deep} or \cite{niroomand2022capacity}. In these reports, loss landscape features are employed to determine whether the substantial cost of using ensembles is justified. Hence, a method to determine ensemble performance \textit{a priori} is highly desirable.
\newline\textbf{Mat\'{e}rn kernel:} The mathematical form of the Mat\'{e}rn kernel is introduced in the following section. Usually, a standard implementation of a half-integer version of the kernel is employed by machine learning practitioners \cite{sklearnpackage}. A helpful account of covariance kernels in general can be found in \cite{genton2001classes}.
For their flexibility, Mat\'{e}rn kernels have also been studied in the context of non-Euclidean, Riemannian geometries \cite{borovitskiy2020matern} and are commonly employed for physical applications due to their reduced smoothness compared to other kernels \cite{prakash2018robust, doctor2017statistical}. A more detailed discussion around kernel decision making is included in Appendix \ref{app:kerneldecmak}.
%In general, theoretical work shows that kernels that rely on a smoothness parameter, which is true for most standard kernels including the Mat\'ern family and RBF kernels, are more prone to issues from the curse of dimensionality \cite{bengio2005curse}. 

\subsection{Contribution and organisation}
In this work, we apply state-of-the-art methods from physics to improve the understanding and interpretability of GPs. To the best of our knowledge, we are the first to study and visualise loss landscapes of the negative log marginal likelihood (-lml) function for GPs. Our interest lies in the analysis of hyperparameter space topology for various standard GP problems and its direct applications towards improving the usability of GPs. In summary, we 
\begin{itemize}
  \item Outline how methods and knowledge from physics can be employed in interpretability for Bayesian machine learning.  
  \item Describe the use of geometric and physical features of the loss landscapes to increase GP ensemble performance and reduce computation time.
  \item Provide novel insights into non-standard kernels by directly including $\nu$ in the optimisation problem and explore the evolving hyperparameter optimisation problem.
\end{itemize}

\section{Notation and Background}
\subsection{Gaussian Process}
Let $f: \mathcal{X} \rightarrow \mathbb{R}$ be a random function on some data $\mathcal{X}$. A Gaussian process (GP) is a collection of random variables $f(x), x \in \mathcal{X}^d$ such that for any finite collection of points and any $d$, the joint distribution of variables in the vector $\mathbf{f(x)}$ is multivariate Gaussian distributed with prior mean vector $\mathbf{\mu}$ and covariance matrix $\mathbf{\Sigma}$. A latent function, $f$, can hence be assigned a GP prior $f \sim \mathcal{GP}(\mu(x),k(x,x^\prime))$, with $\mu(x) = 0$ being a common choice for the mean function and $k(x,x^\prime)$ being some covariance kernel. Thus, the latent function in a GP is represented by samples from an infinite multivariate Gaussian distribution, parameterised by a mean and covariance function. Given the GP prior $f$, together with some training data $\mathcal{D} = \{\mathbf{\mathcal{X}}, \mathbf{y}\}$, and allowing for noise by setting $y_i = f(x_i)+\epsilon_i,\ \epsilon_i \sim \mathcal{N}(0,\sigma^2)$, the posterior $p(f|\mathbf{y})$ is conveniently also Gaussian distributed with
\begin{equation}
\begin{aligned}
& \mu(\mathbf{x_\ast}) = \mathbf{\Sigma}_{\mathbf{x}_\ast,\mathbf{x}}(\mathbf{\Sigma}_{\mathbf{x},\mathbf{x}}+\sigma^2\mathbf{\mathrm{I})^{-1}}\mathbf{y}\\
& \sigma^2(\mathbf{x_\ast}) = \mathbf{\Sigma}_{\mathbf{x_\ast},\mathbf{x_\ast}}-\mathbf{\Sigma}_{\mathbf{x_\ast},\mathbf{x}}(\mathbf{\Sigma}_{\mathbf{x},\mathbf{x}}+\sigma^2\mathbf{\mathrm{I})^{-1}}\mathbf{\Sigma}_{\mathbf{x},\mathbf{x_\ast}}
\end{aligned}
\end{equation}
for some new testing point $\mathbf{x_\ast}$, where $\mathrm{I}$ is the identity matrix. By virtue of the Bayesian approach, the posterior can be updated for incoming data, with the values of $\mu(\mathbf{x_\ast})$ specifying the prediction and $\sigma^2(\mathbf{x_\ast})$ the predictive uncertainty (variance) at point $\mathbf{x_\ast}$. In training a GP, we commonly minimise the negative of the log marginal likelihood function
\begin{equation}
    \log p(\textbf{y}|\mathbf{\mathcal{X}}, \theta) =  - \frac{1}{2}\textbf{y}^\intercal \mathbf{\Sigma}^{-1}\textbf{y}-\frac{1}{2} \log|\mathbf{\Sigma}|-\frac{n}{2}\log2\pi,
\end{equation} 
for a set of hyperparameters in hyperparameter space $\theta \in \Theta$, which defines the functional space of the loss landscape. The -lml is a suitable objective function, since it contains both a model complexity term and a data fit term which can counterbalance each other for a well-performing model.

\subsection{Mat\'ern kernel}
One of the most important decisions when designing a Gaussian process is the choice of kernel or covariance function, which computes the correlation (covariance) between any pair of points in the training dataset $\mathcal{D}$. Element $k_{ij}$ of a covariance matrix $\mathbf{\Sigma}$ contains the covariance between points $i$ and $j$. 
%The Mat\'ern kernel is an extension of the popular radial basis function (RBF) kernel by an additional smoothness parameter $\nu$.
The general Mat\'ern kernel is defined as
\begin{equation}
k_{ij} = k\left(x_{i}, x_{j}\right)=\sigma^2\frac{1}{\Gamma(\nu) 2^{\nu-1}}\left(\frac{\sqrt{2 \nu}}{\ell} d\left(x_{i}, x_{j}\right)\right)^{\nu} K_{\nu}\left(\frac{\sqrt{2 \nu}}{\ell} d\left(x_{i}, x_{j}\right)\right)
\end{equation}
where $\Gamma(\cdot)$ is the Gamma function, $K_{\nu}$ is the modified Bessel function of the second kind of order $\nu$ and $d(\cdot,\cdot)$ is the Euclidean distance between two points. The kernel is parameterised by an amplitude $\sigma$ and a lengthscale $\ell^d$, which for the anisotropic kernel we use here is a vector of length $d$ for input data $x \in \mathbb{R}^d$. %We have also considered a single lengthscale kernel, but the anisotropic version outperforms the single lengthscale kernel due to its increased flexibility.

While the Mat\'ern function can theoretically be evaluated for all values of $\nu$, for which it is $\lceil \nu \rceil -1$ differentiable, there exist special solutions at half integers. The commonly used 1/2, 3/2 and 5/2 Mat\'ern kernels are realised as
\begin{equation}
\begin{aligned}
&M_{1 / 2}(d)=\sigma^{2} \exp \left(-\frac{d\left(x_{i}, x_{j}\right)}{\ell}\right), \\
&M_{3 / 2}(d)=\sigma^{2}\left(1+\frac{\sqrt{3} d\left(x_{i}, x_{j}\right)}{\ell}\right) \exp \left(-\frac{\sqrt{3} d}{\ell}\right), \\
&M_{5 / 2}(d)=\sigma^{2}\left(1+\frac{\sqrt{5} d\left(x_{i}, x_{j}\right)}{\ell}+\frac{5 d\left(x_{i}, x_{j}\right)^{2}}{3 \ell^{2}}\right) \exp \left(-\frac{\sqrt{5} d\left(x_{i}, x_{j}\right)}{\ell}\right).
\end{aligned}
\end{equation}
The Mat\'ern $M_{1/2}$ kernel is identical to the absolute exponential kernel. The smoothness of the Mat\'ern kernel increases with $\nu$, and for $\nu \rightarrow \infty$, it becomes equivalent to the RBF kernel. In practice, low values of $\nu$ may be chosen because the reduced smoothness can fit complex data more accurately. Derivatives of the Mat\'ern kernel with respect to its smoothness parameter $\nu$ require derivatives of the modified Bessel function of the second kind $K_\nu$ w.r.t. $\nu$, for which numerical derivatives are usually employed. %We have developed a Fortran implementation of Bessel functions and its derivatives to compute non half-integer $\nu$ parameterisations, since an implementation in Python was too slow.
To increase efficiency and accuracy of these derivative computations, \textcite{geoga2022fitting} have recently presented an implementation using automatic differentiation. However, computing non-half integer kernels remains expensive and a bottleneck in motivating the use of such values of $\nu$.

\section{Empirical observations}
To show the value of characterising the loss landscape using methods from physics, we present selected examples of how these methods can improve our understanding of GPs. The examples form a non-exhaustive list and we note that an extension to other kernels, to which these methods generalise without any restrictions, is a natural avenue for future work. Here, we present results that focus on hyperparameter choices for the Mat\'ern kernel and loss landscape guided GP ensembles. 

\subsection{\texorpdfstring{Mat\'ern kernel for changing $\nu$}{Matern kernel for changing Nu}}
Characterising the -lml loss landscape, parameterised by the GP hyperparameters, as a function of different values for Mat\'ern-$\nu$, requires sampling significant regions of the loss landscape. Thus, we reconstruct the loss landscape using basin-hopping global optimisation \cite{lis87,walesd97a,waless99} for each value of $\nu$ in the interval $[0.5,4.0;0.1]$. Basin-hopping optimisation involves taking a jump to different region of the landscape once a minimum is identified. After local (L-BFGS) minimisation, this jump is accepted under a Metropolis criterion of $P \propto \exp (-\Delta \mathcal{L}/c)$ for some constant $c$ and a difference in loss value between the old minimum and the new point $\Delta \mathcal{L}$. If accepted, the local minimum becomes the current minimum and the process resumes. Due to the theoretical unboundedness of the GP loss landscape space, the algorithm stops after a convergence criterion is reached. Such a criterion may be a fixed number of steps or more commonly be defined by some number $n$ for which no new minima were identified in the last $n$ basin-hopping steps. Further details on landscape exploration methodologies are provided in Appendix \ref{sec:appc}.

\begin{figure}[!h]
\centering
  \begin{subfigure}[t]{0.03\linewidth}
    (a)
  \end{subfigure}
  \begin{subfigure}[t]{0.45\linewidth}
    \includegraphics[width=\linewidth, valign=t]{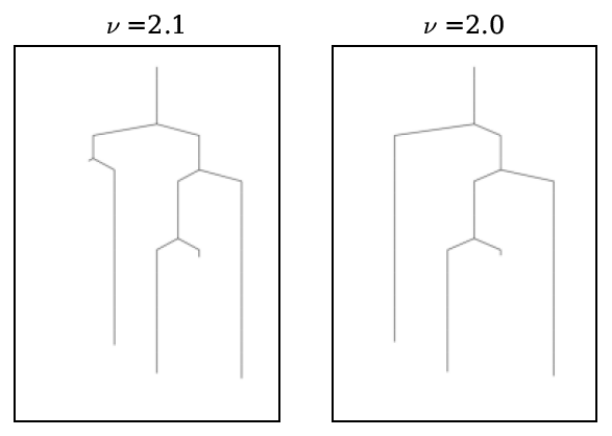}
  \end{subfigure}\hfill
  \begin{subfigure}[t]{0.03\linewidth}
    (b)
  \end{subfigure}
  \begin{subfigure}[t]{0.45\linewidth}
    \includegraphics[width=\linewidth, valign=t]{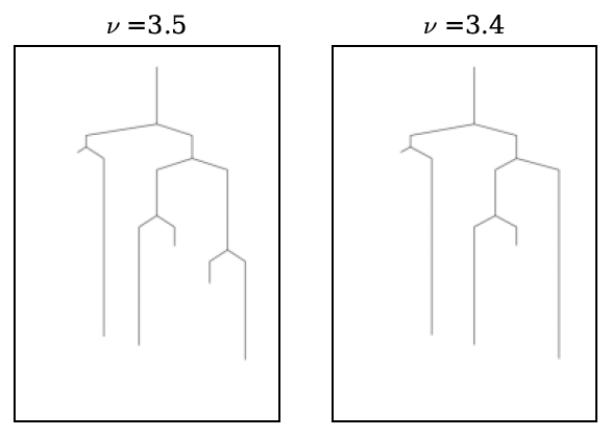}
  \end{subfigure}
\caption{Loss landscape fold catastrophes illustrated in disconnectivity graphs. Each leaf node of the graphs is a minimum in hyperparameter space $\Theta$. The disappearance of  minima from $\nu{=}2.1 \rightarrow \nu{=}2.0$ (a) and $\nu{=}3.5 \rightarrow \nu{=}3.4$ (b) corresponds to a reduction in the number of leaf nodes.}
\label{fig:fold_catsts}
\end{figure}

We begin by considering the Schwefel function, $d-$dimensional function defined by
\begin{equation}
    f(x) = 418.9829d- \sum_{i=1}^dx_i\sin{\sqrt{|x_i|}}.
\end{equation}
The Schwefel function is commonly used to study GP optimisation problems and provides a way to test GPs on a complex, well-defined problem. A more detailed perspective of the Schwefel function and the corresponding GP fits is included in Appendix \ref{app:schwefel}. Figure \ref{fig:minima_loss_mse} shows the loss value of all minima for fitting 400 datapoints in the [-100,100] hypercube of a 3$d$ Schwefel function with various values of $\nu$. We observe a continuous, smooth change in loss value of minima as a function of increasing $\nu$. The clustering highlights which minima are qualitatively the same, and are preserved across all landscapes. Most minima are present at almost the entire $\nu$-range. The loss landscapes for sequential values of $\nu$, visualised as disconnectivity graphs (Fig. \ref{fig:dgs} in Appendix \ref{sec:appc}) underlines the $\nu-$dependent continuity. Changes in the landscapes are largely continuous and smooth, with the exception of minima that disappear at $\nu \leq 2.1$ and $\nu \leq 3.5$ as shown in figure \ref{fig:fold_catsts}. In both cases, an additional minimum exists for higher $\nu$ but disappears when $\nu$ decreases below the threshold. In figure \ref{fig:fold_catsts}, the loss landscape is represented by disconnectivity graphs in a coarse-grained fashion, characterised only by minima as leaf nodes and saddle points that act as transition states between minima, as parent nodes. A more detailed discussion of disconnectivity graphs and further examples are given in Appendix \ref{sec:appc}.

As $\nu$ decreases towards the critical points at which we observe a disappearance of minima, the barriers separating the disappearing minimum with an adjacent one smoothly decrease. This reduction indicates minimum removal through a process analogous to a fold catastrophe in statistical physics, in which a transition state and minimum coalesce. Catastrophe theory dictates this behaviour for a continuous parameter change such as that for $\nu$ \cite{zeeman1976catastrophe}. Previously, \textcite{leonardos2022exploration} have considered fold catastrophes for machine learning loss landscapes in Q-learning, which we extend here to nonparametric machine learning and GPs. These non-smooth, radical transitions between subsequent values of $\nu$ highlight the importance of hyperparameter tuning in GPs, and specifically the importance of non-standard (half-integer) values of $\nu$. Minima that suddenly disappear in a fold catastrophe are a concern to the practitioner who may observe large discontinuous changes in model performance for small perturbations in $\nu$. Additional experiments to confirm that, outside the fold catastrophes highlighted above, individual values of $\nu$ do have smoothly transitioning solution spaces are included in Appendix \ref{sec:appa}. 
\newline\newline
In studying the effects of changing $\nu$, we decided to focus on the range $[0.5,4.0]$ since it encompasses all the commonly used Mat\'ern kernels $\nu \in \{0.5,1.5,2.5\}$ and also explores the regime of larger $\nu$ values. In figure \ref{fig:minima_loss_mse}, we compare the loss values of all minima to the corresponding mean squared error (MSE). The emerging pattern is identical for the two plots, and the sequential changes between different values of $\nu$ can be seen easily. Minor differences are observed, which is expected, since loss does not necessarily correlate perfectly with test performance, due to differences in the distribution of points in training and testing datasets (figure \ref{fig:minima_loss_mse}).

\begin{figure}[!h]
\centering
  \begin{subfigure}[t]{0.03\linewidth}
    (a)
  \end{subfigure}
  \begin{subfigure}[t]{0.45\linewidth}
    \includegraphics[width=\linewidth, valign=t]{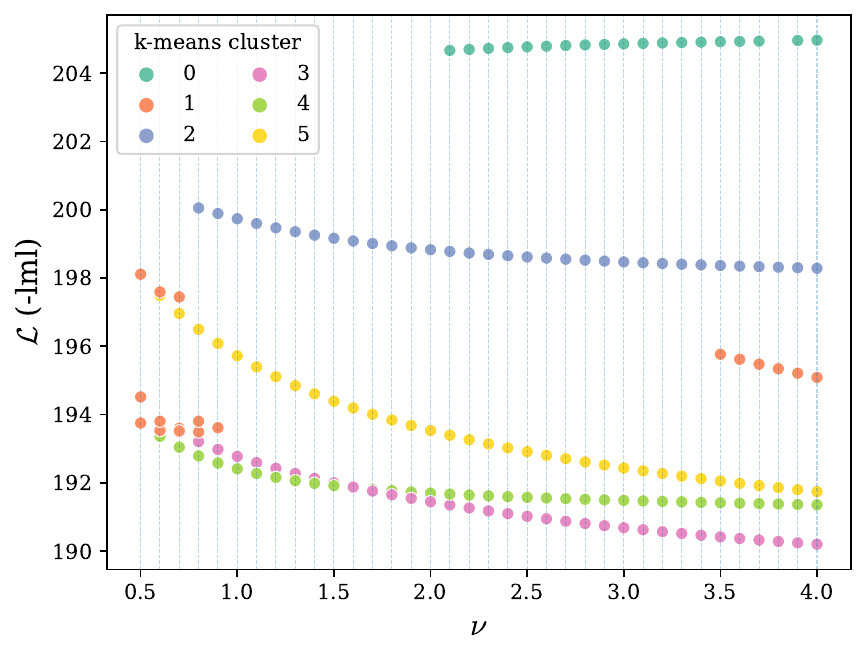}
  \end{subfigure}\hfill
  \begin{subfigure}[t]{0.03\linewidth}
    (b)
  \end{subfigure}
  \begin{subfigure}[t]{0.45\linewidth}
    \includegraphics[width=\linewidth, valign=t]{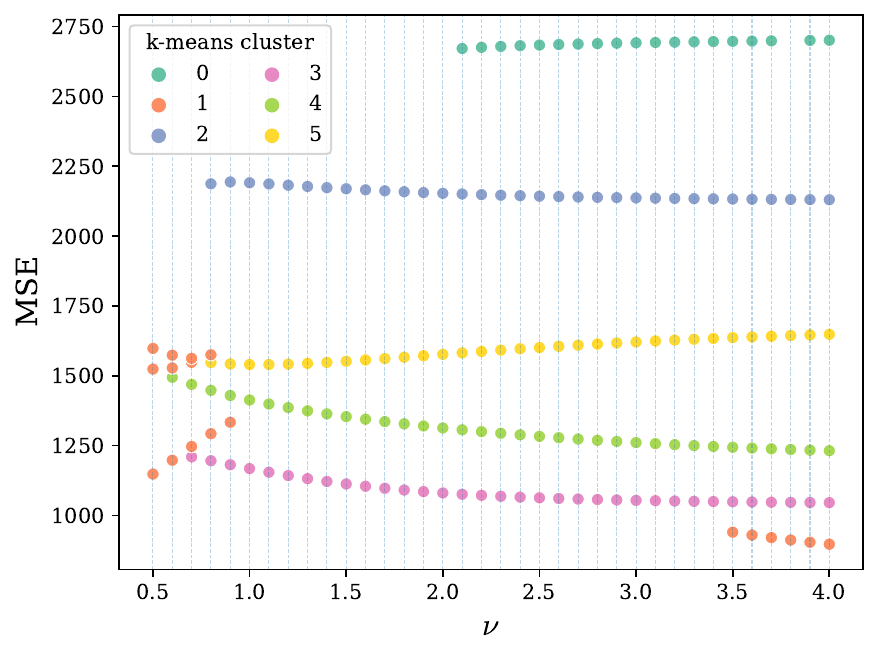}
  \end{subfigure}
   \caption{Loss value (-lml) (a) and mean squared error (b) for all identified minima at all values of $\nu$. $k$-means clustering is performed on the minima, where $k$ is chosen to be the maximum number of minima at any $\nu$. The data shown here were obtained from the 3$d$ Schwefel function.}
   \label{fig:minima_loss_mse}
\end{figure}

\subsection{\texorpdfstring{$\nu$ in hyperparameter optimisation}{Nu in hyperparameter optimisation}}
To better understand the role that the smoothness parameter, $\nu$, plays in the Mat\'ern kernel, we include it in the hyperparameter optimisation. Instead of fixing $\nu$ and optimising the -lml surface for the remaining hyperparameters, we optimise $\nu$ as part of $\theta$. The lack of previous work on this idea can perhaps largely be attributed to the increase in computational cost of computing non-half-integer solutions and the absence of this method in standard Python packages. One of the few works that study non-half-integer values of $\nu$ is the study of the bandit problem in reproducing kernel Hilbert spaces (RKHS) by \textcite{janz2020bandit}. 
\newline\newline
For the 3-dimensional Schwefel function and a varying number of datapoints with repeated sampling, we obtain the results shown in table \ref{tab:optnu}. In general, the best test MSE improves substantially with an increased number of training points, whilst the training MSE is highest for a small number of datapoints, clearly indicating overfitting in the low-data regimes. The main result, however, is that the best value for $\nu$ stabilises for increasing $n$ in the region of $\nu=4.6$. Importantly, when we compare the $n=900$ case with a standard $\nu=2.5$ kernel, we find that including $\nu$ in the optimisation performs around 18\% better, since the MSE of a standard Mat\'ern $5/2$ kernel is 2.8 for the same dataset. Hence, we provide empirical proof that the commonly used parameterisations of $\nu$ are far from optimal in terms of inference accuracy in various scenarios. These results transform the discussion around kernel choice into a standard speed-accuracy trade-off problem. Picking half-integer $\nu$ values is substantially faster, yet may also be far less accurate.

\begin{table}[!ht]
\caption{GP performance on the 3$d$ Schwefel dataset with $\nu$ included in the hyperparameter space $\Theta$. $n$ is the number of datapoints, $t$ the CPU (12 core Intel X5650) time, in hours, to complete landscape exploration. 
%Using GPUs, the process can be parallelised and hence substantially accelerated. 
The MSE is the best MSE of all identified minima, the corresponding $\nu$ is the $\nu$-value of the minimum with the best MSE. }
\vspace{0.3cm}
  \centering
  \begin{tabular}{l{c}{c}{c}{c}{c}{c}{c}l}
    \toprule
    \multicolumn{3}{c}{\phantom{a}} & \multicolumn{2}{c}{test} & \multicolumn{2}{c}{train}  \\
    \cmidrule(r){4-5}
    \cmidrule(r){6-7}
    $n$ & \#minima &  $t$ & MSE & $\nu$ & MSE & $\nu$  \\
    \midrule
    100 & 42 & 0.8 & 280.7 & 2.29 & 1.9e-10 & 1.01 \\
    200 & 14 & 9.3 & 125.1 & 5.15 & 1.1e-08 & 5.15 \\
    300 & 10 & 10.1 & 43.5 & 1.55 & 1.6e-09 & 3.53 \\
    400 & 10 & 18.5 & 28.3 & 4.83 & 1.9e-08 & 4.83 \\
    500 & 15 & 35.2 & 19.1 & 3.77 & 1.8e-08 & 3.77 \\
    600 & 10 & 67 & 7.7  & 4.74 & 4.1e-08 & 4.74 \\
    700 & 7 & 79 & 6.9 & 4.34 & 7.0e-08 &  4.34 \\
    800 & 8 & 94.2 & 2.8 & 4.67 & 9.1e-08 & 4.67 \\
    900 & 13 & 176.1 & 2.3 & 4.45 & 1.2e-07 & 4.45 \\
    1000 & 7 & 181.2 & \textbf{2.6} & 4.87 & 2.7e-07 & 4.87 \\
    \bottomrule
  \end{tabular}
  \label{tab:optnu}
\end{table}

\subsection{Physics-inspired ensemble learning}
Combining the outputs of multiple predictors has become common practice in machine learning. Instead of relying on a single model, multiple ML models perform inference on the same task and a weighted aggregation of these individual inferences forms the final prediction. In practice, multiple predictive models are usually identified by repeated stochastic gradient descent beginning from new random initialisations of the model. However, applications of GP ensembles remain sparse, both due to computational cost and insufficient model performance. We provide a novel ensembling approach that aims to tackle both issues. Using the energy landscapes methodology from theoretical molecular sciences \cite{wales2003energy}, we are able to identify multiple distinct minima of the loss landscape, similar to the work done by \textcite{niroomand2022capacity} for neural networks. For the purpose of GP ensembles, each distinct minimum can be viewed as an individual set of hyperparameters to characterise a single model. Thus, in contrast to the classical ensemble generation process, this landscape inspired methodology guarantees unique minima without the risk of including the same minimum multiple times. By employing methods from the field of energy landscapes, we exploit knowledge and insights from a completely different area of science to construct GP ensemble models in an entirely novel way. Over various experiments, we show that these ensembles substantially outperform any single minimum of the same loss landscape by a significant margin for various synthetic and real-life datasets. We furthermore show that the choice of voting or weighting scheme between individual classifiers is extremely important and greatly impacts model accuracy.

\subsubsection{Ensemble methodology and performance}
\begin{figure}[!htb]
\centering
  \begin{subfigure}[t]{0.03\linewidth}
    (a)
  \end{subfigure}
  \begin{subfigure}[t]{0.45\linewidth}
    \includegraphics[width=\linewidth, valign=t]{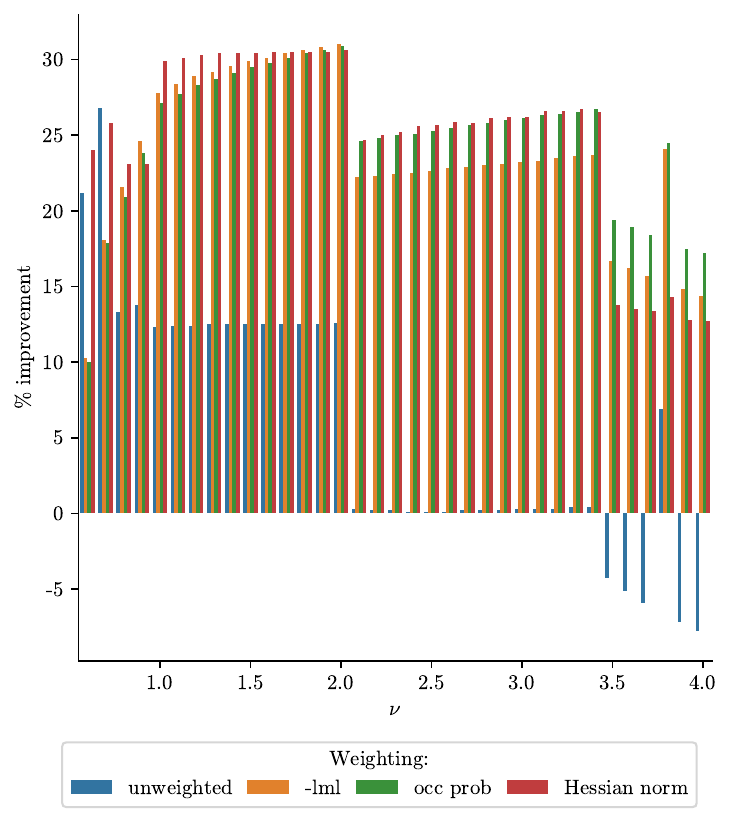}
  \end{subfigure}\hfill
  \begin{subfigure}[t]{0.03\linewidth}
    (b)
  \end{subfigure}
  \begin{subfigure}[t]{0.45\linewidth}
    \includegraphics[width=\linewidth, valign=t]{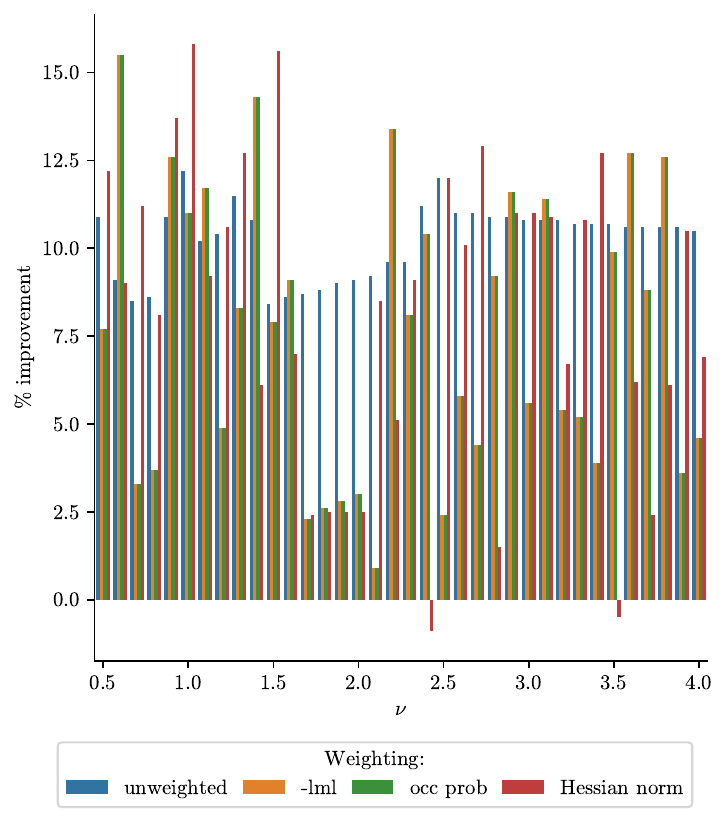}
  \end{subfigure}
   \caption{Percentage improvement over single best minimum of GP ensembles for fitting the (a) 3$d$  and (b) 4$d$ Schwefel function for various parameterisations of $\nu$. Different weighting strategies are highlighted in respective colours.}
   \label{fig:ensemble_comparison}
\end{figure}

Combinations of the individual predictions from all considered minima can be performed by simple majority vote or uniform-weight based prediction across all considered minima. However, we have also tested various other weighting schemes, inspired by common methods in the physical sciences. Results are shown in Table \ref{tab:optnu}. The three weighting schemes considered are weighting by loss value, occupation probability, and spectral norm of the Hessian. 
\newline\textbf{Loss value:} Weighting ensemble minima by loss value makes the prediction a function of performance on the training set. We also considered weighting by inverse training MSE, but did not observe a visible difference between these schemes. Weighting by loss value performs very well in all datasets, as shown in table \ref{tab:best_ensembles}. This performance highlights both the value of the best minimum in the loss landscape and the fact that evidently, additional information contained in other minima contributes to an improvement during inference.
\newline\textbf{Occupation probability:}
The occupation probability is a function of the log product of positive Hessian eigenvalues at a given local minimum. A detailed account of how the occupation probability is computed is included in Appendix \ref{app_occprob}. The occupation probability provides an approximate of the width or flatness of a minimum. Minimum flatness is often seen as a proxy for robustness to perturbations in the input data \cite{hochreiter1997flat, hinton1993keeping, zhang2018energy}. A higher occupation probability may therefore indicate a wider and more robust minimum. Thus, weighting by occupation probability is expected to give more robust minima a higher weight, leading to better test MSE. We can see significant improvements in test MSE for this weighting method in table \ref{tab:best_ensembles}.
\newline\textbf{Spectral norm of Hessian:} The Hessian norm at a minimum provides a geometric interpretation of flatness, which, like the occupation probability, is expected to correlate with robustness. Hence, it is no surprise that occupation probability and Hessian norm perform similarly, with the Hessian performing slightly better on some datasets.
\newline\newline
The performance of GP ensembles for different values of $\nu$ for 3- and 4-dimensional Schwefel data are shown in figure \ref{fig:ensemble_comparison}. As above, a largely smooth progression across values of $\nu$ is observed, more so in the less complex 3$d$ Schwefel problem. It is noteworthy that ensembles outperform the single best minimum in almost all cases, and that landscape-guided weighting schemes almost always outperform simple majority vote combinations. This result strengthens the argument for using carefully crafted GP ensembles, particularly when designed under the consideration of loss landscape features.

Table \ref{tab:best_ensembles} describes the best performance of each ensemble method for various datasets and shows how computing geometric features of the loss landscape can help to improve the model at inference time. The QSAR biodegradation dataset \cite{mansouri2013quantitative} is a collection of 1055 chemical molecules characterised by 41 molecular descriptor attributes with the aim of predicting the bioconcentration factors (BCF), which here serves as a proxy for toxicity. For each of the datasets, both synthetic and from the real world, weighting the minima by any landscape feature outperforms the unweighted combination by around 20-25\%.

\begin{table}[h!]
\caption{MSE improvement of GP ensemble as percentage compared to single best minimum from global optimisation. Improvements for each of the considered weighting schemes are shown with best performing scheme highlighted.}
\vspace{0.3cm}
  \centering
  \begin{tabular}{l{c}{c}{c}{c}{c}{c}{c}l}
    \toprule
     & Schwefel 3$d$ &  Schwefel 4$d$ & QSAR  \\
    \midrule
    Unweighted & 26.8 & 12.3 & -12.6 \\
    -lml & \textbf{31.1} & 15.5 & 19.1 \\
    Occupation probability & 30.9 & 15.5 & 11.0 \\
    Spectral norm of Hessian & 30.6 & \textbf{16.1} & \textbf{22.9} \\
    \bottomrule
  \end{tabular}
  \label{tab:best_ensembles}
\end{table}

\subsubsection{To ensemble or not to ensemble}
\begin{figure}[!htb]
\begin{tabular}{cc}
    \begin{minipage}{0.48\textwidth} 
    \includegraphics[width=0.9\textwidth]{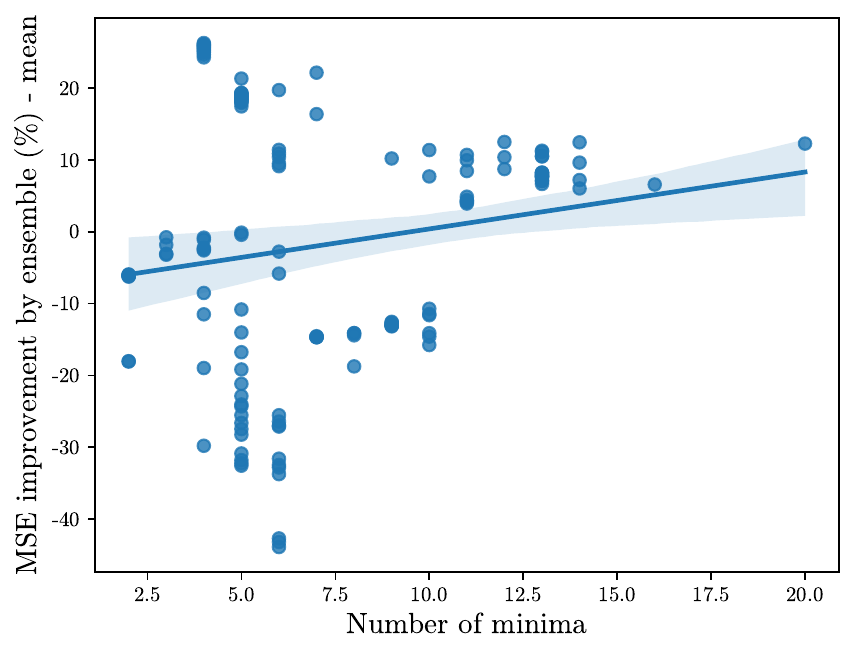} \\ \includegraphics[width=0.9\textwidth]{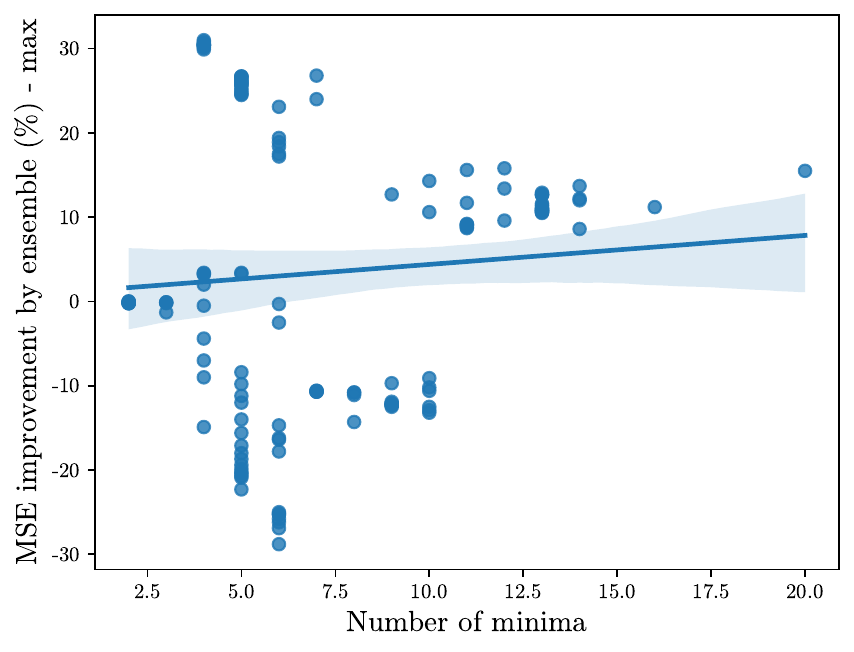} \end{minipage} & \hspace{-1cm} 
    \begin{minipage}{0.49\textwidth} \includegraphics[width=1.15\textwidth]{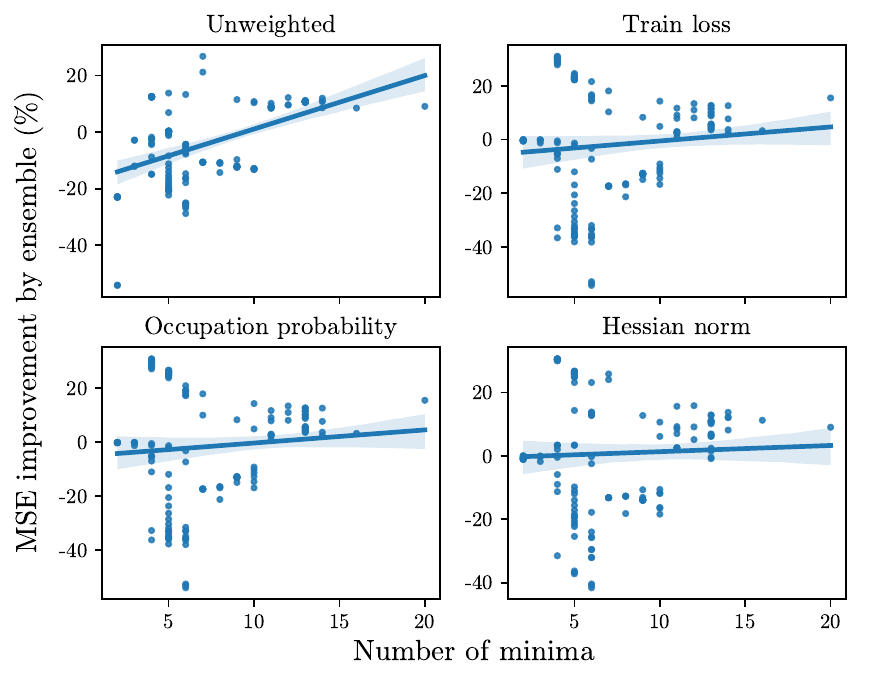} \end{minipage}
    \end{tabular}
        \caption{Correlation between MSE improvement achieved by ensemble methods and the number of minima in a loss landscape. The underlying data is aggregated over multiple regression problems, both real-world and synthetic. \label{fig:nmin_vs_ensemble}}
    \end{figure}

The decision to use an ensemble of models in a given ML problem involves a trade-off between accuracy and computational cost. Ensembles often improve inference, but require substantial additional computational resources, especially if a large number of individual predictors are queried. Estimating the value of an ensemble, relative to a single model, prior to construction, will allow better resource allocation. Hence, a means of \textit{a priori} estimating whether ensembles are beneficial would be useful. Here, we present and test an empirical method to estimate the benefit of using ensembles by considering features of the loss landscape. 
\newline\newline Figure \ref{fig:nmin_vs_ensemble} reveals a positive correlation between MSE improvement by ensembles and the number of minima in the loss landscape. When the number of minima is small, in our cases ${<1}0$, ensembles perform worse than the single best minimum. Some minima provide poor solutions that corrupt the weighted average in the ensemble generation process.
The small numbers of minima in GPs compared to neural networks \cite{niroomand2022characterising, verpoort2020archetypal} can, as discussed above, be largely attributed to the lower dimensionality of the optimisation space. In GPs, only a handful of hyperparameters are optimised when minimising the -lml, as opposed to millions in large neural networks. 
\newline\newline
The plots in figure \ref{fig:nmin_vs_ensemble}  showing weighting-strategy correlation provide further evidence that ensembles require a large number of minima to average out bad predictors. All weighting schemes have low correlation, indicating that they successfully manage to select against bad minima, while the unweighted ensemble exhibits extremely strong correlation. Therefore, GP ensembles using simple majority voting are usually insufficient for models that are concurrent with small numbers of minima in the loss landscape. This result also strengthens the argument for landscape-based weighting schemes in GP ensembles, such as those shown in figure \ref{fig:ensemble_comparison}. These more elaborate weighting schemes discriminate against bad minima and hence are vital for accurate inference.

\section{Conclusions and future work}
Understanding the loss landscapes of Gaussian processes has received surprisingly little attention, which is likely due to the lack of readily available methodology and high computational cost. We have demonstrated that knowledge about the loss landscape structure can improve Gaussian process model performance, provide increased confidence in the trained model, and decrease computational cost. Naturally, computing large areas of the loss landscape to sample minima comes at a cost, but this cost is relatively low compared to extensively sampling and computing solutions for various minima in common ensemble methods. However, a speed-accuracy/explainability trade-off is inherent to the loss landscape approach for studying GPs. The same trade-off occurs when considering non-standard (half integer) parameterisations of the Mat\'ern kernel, which may lead to substantially lower inference errors at higher computational cost. Lastly, we have provided empirical evidence that transitions between adjacent values of $\nu$ in Mat\'ern kernels occur smoothly, which provides a novel insight into the structure of the solution space for GPs. 
\newline\newline
An exciting area of future work in GP loss landscapes will be improved hyperparameter sampling methods based on geometric properties of the loss landscape using methods from physics. A fully Bayesian treatment of a GP would imply the elimination of single point estimates for hyperparameters in favour of a distribution of the hyperparameters to sample from. Currently, this sampling is often performed using Hamiltonian Monte Carlo (HMC) \cite{saatcci2010gaussian} methods, which are computationally expensive.
Such HMC methods could potentially be replaced by a landscape approximation constructed from the Hessian around minima, and transition state information. This approximated landscape could be used as the inverse sampling distribution for a Bayesian treatment of hyperparameter sampling at a substantially lower cost compared to prevalent methods. 

\newpage
\printbibliography
\newpage
\appendix
\section{Appendix}
\renewcommand{\thesubsection}{\Alph{subsection}}
\renewcommand\thefigure{A.\arabic{figure}}    
\setcounter{figure}{0}

\subsection{Out of distribution generalisation for Schwefel function}\label{app:schwefel}
\begin{figure}[!h]
\centering
   \begin{subfigure}{\linewidth}
   \includegraphics[width=1\textwidth]{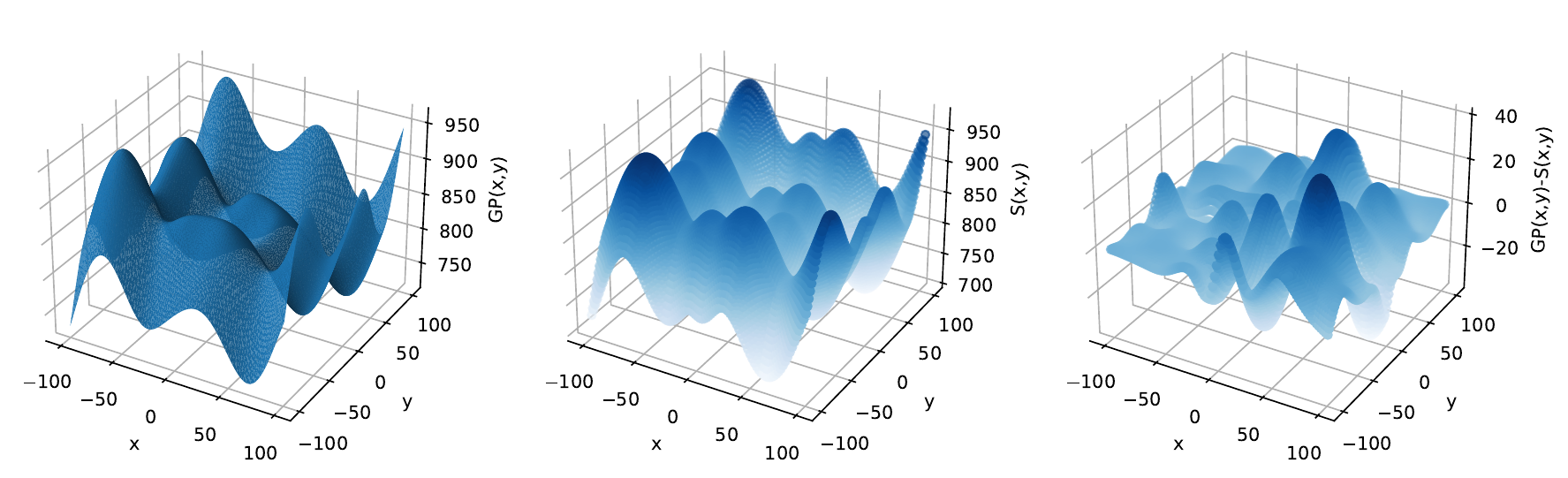}
\end{subfigure}
\begin{subfigure}{\linewidth}
   \includegraphics[width=1\textwidth]{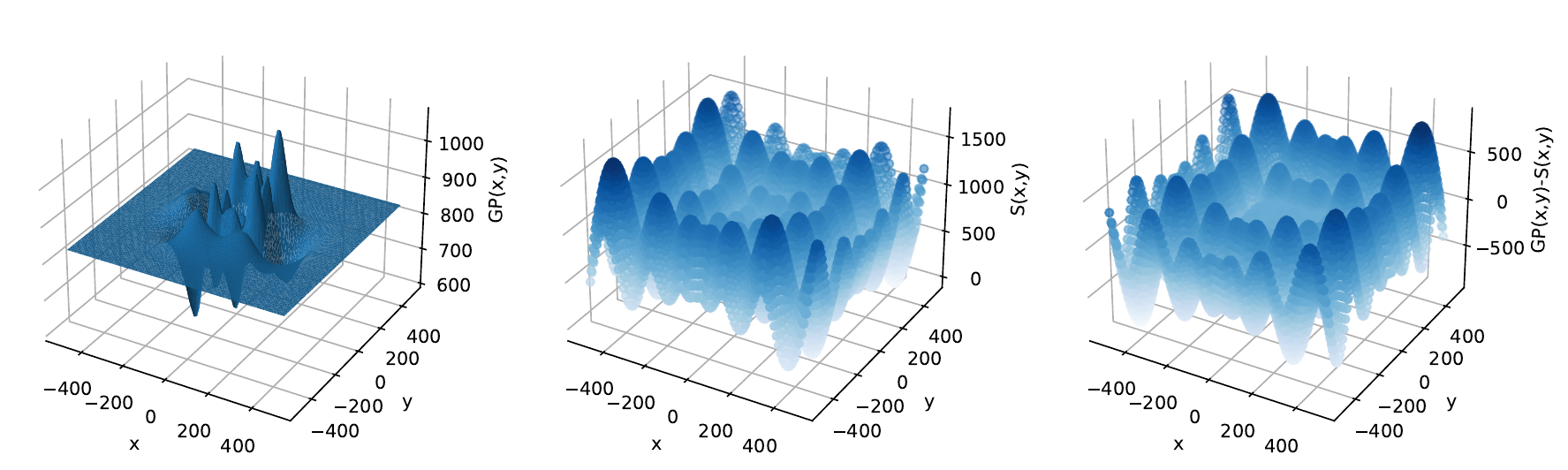}
\end{subfigure}
\caption{Gaussian Process on a 2D realisation of the Schwefel problem. Model trained on the [-100,100] hypercube. Top figure shows from left to right 1) GP predictions, 2) true Schwefel function, 3) predicted-true landscape. Bottom panels show the function over a larger domain, still trained on [-100,100] hypercube.}
   \label{fig:schwefel}
\end{figure}

To study loss landscapes for GPs, we begin by considering the Schwefel function, a non-convex, $d-$dimensional function defined by
The Schwefel function is commonly used to study GP optimisation problems and provides a way to test GPs on a complex, well-defined problem. A visualisation of the Schwefel function is given in  Figure \ref{fig:schwefel}. The GP model performs very well in the region it has been trained on, yet generalises poorly to out-of distribution samples far away from training data. The flatness of the surface at regions far away from training data is typical of Bayesian kernel methods. Since the GP has no knowledge of this region, and since no information is explicitly encoded into the kernel as prior information, the predicted function in this range is simply flat.   

\subsection{Kernel decision making}\label{app:kerneldecmak}
The choice of an appropriate kernel for a given dataset has been the subject of active research, mainly focusing on the prior knowledge of the data practitioner making the kernel choice.
\textcite{giordano2022evaluating} considers uncertainty in the human prior and studies the robustness and effect of varying that prior on outcome prediction. 
In contrast, \textcite{stephenson2021measuring} consider qualitatively interchangeable kernels, namely kernels that, given a prior, should both lead to the same result, and show that outcome invariance is not necessarily obeyed. 
To improve kernel decision making, others have explored automatic kernel selection \cite{benton2019function, wilson2016deep, abdessalem2017automatic}. However, automatic kernel selection is inherently limited by prior knowledge, and large parts of kernel space remain unexplored. Currently, practitioners choose from a handful of readily available kernels, and disregard other options.

\subsection{\texorpdfstring{Landscape disconnectivity graphs for varying $\nu$}{Landscape disconnectivity graphs for varying Nu}}\label{sec:appc}
The continuous, smooth changes in the loss landscape with changing $\nu$ are characterised by smooth changes of minima in figure \ref{fig:minima_loss_mse}. This view is strongly reinforced by the disconnectivity graphs \cite{wales1998archetypal, krivov2002free}, which provide a coarse-grained, low-dimensional representation of the landscape, characterised by minima and transition states. Disconnectivity graphs can be computed using the disconnectionDPS subroutine \cite{disconnectionDPS} which is available under GNU GPL. Each terminal leaf node in a disconnectivity graph is a minimum, and each parent node represents a transition state between minima. Transition states are index-1 saddle points which act as energy barriers between any 2 adjacent minima. Various methods for finding transitions states are discussed in \cite{wales2003energy} including doubly-nudged elastic band (DNEB) and hybrid Eigenvector following methods. In disconnectivity graphs, ordering along the horizontal axis is arbitrary and the vertical axis represents the loss value, here -lml. The disconnectivity graphs for adjacent values of $\nu$ look extremely similar, with slight changes in the loss values.

\begin{figure}[!h]
    \centering
    \includegraphics[width=0.8\textwidth]{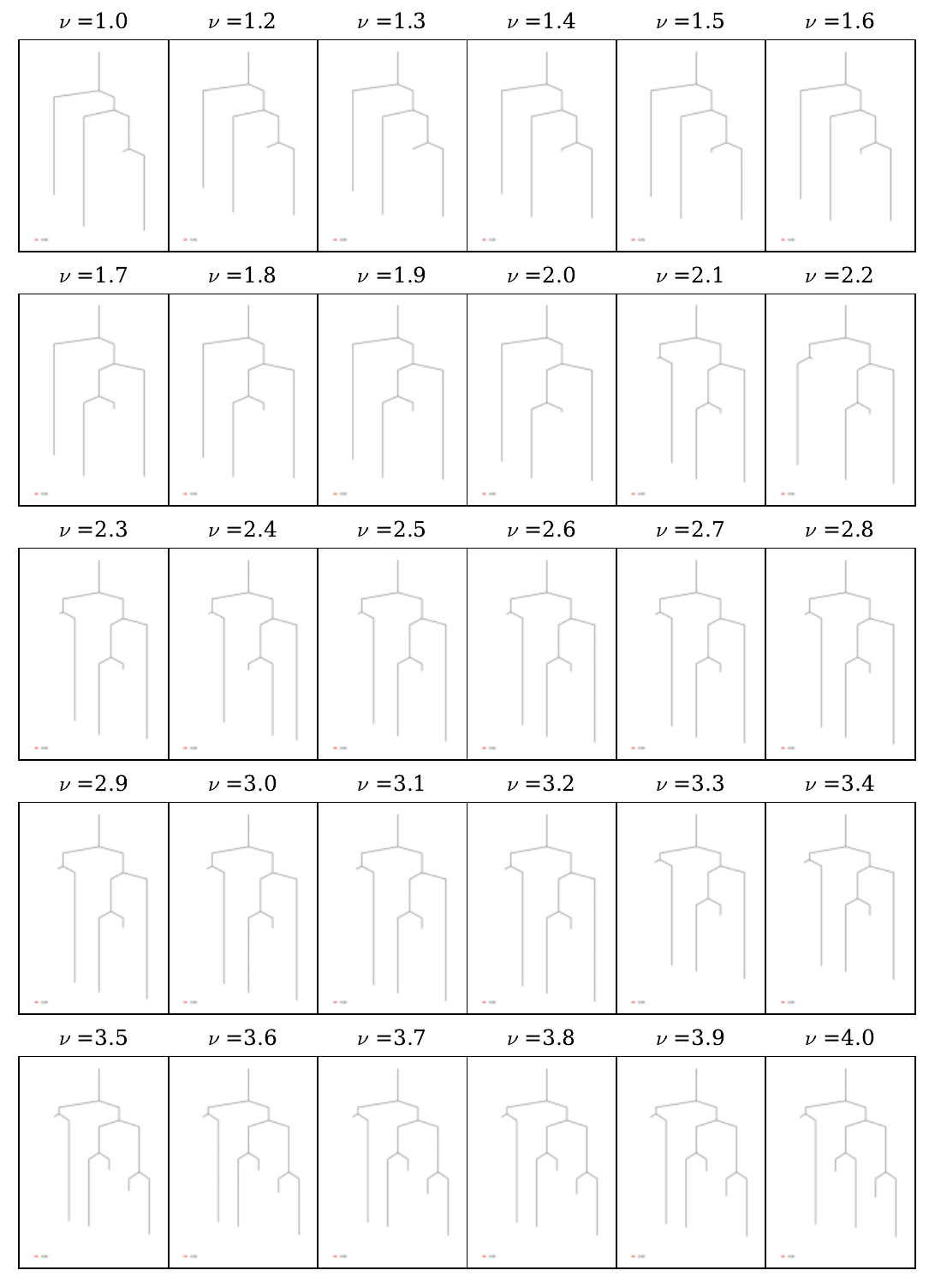}
    \caption{Disconnectivity graphs visualising loss function landscapes for sequential values of $\nu$ when fitting the 3$d$ Schwefel function with $n$ points.}
    \label{fig:dgs}
\end{figure}

\subsection{\texorpdfstring{Minima for varying $\nu$}{Minima for varying Nu}}\label{sec:appa}
\begin{figure}[!h]
\centering
  \begin{subfigure}[t]{0.03\linewidth}
    (a)
  \end{subfigure}
  \begin{subfigure}[t]{0.45\linewidth}
    \includegraphics[width=\linewidth, valign=t]{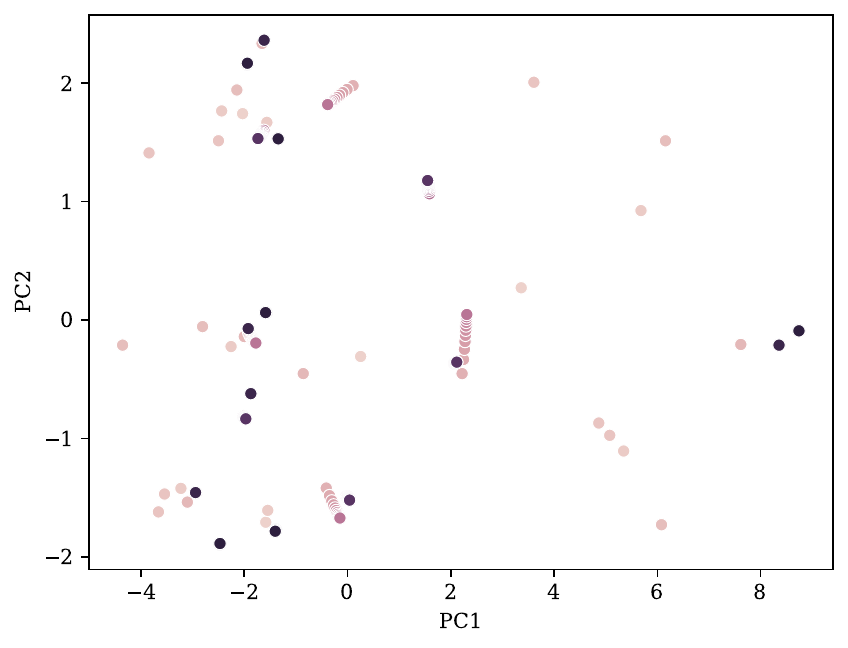}
  \end{subfigure}\hfill
  \begin{subfigure}[t]{0.03\linewidth}
    (b)
  \end{subfigure}
  \begin{subfigure}[t]{0.45\linewidth}
    \includegraphics[width=\linewidth, valign=t]{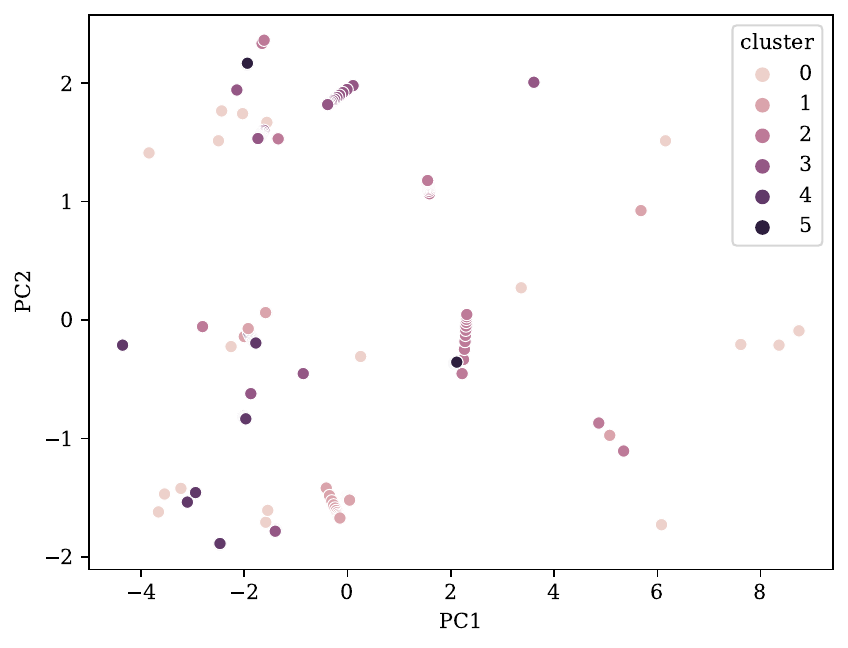}
  \end{subfigure}
   \caption{PCA of individual minima for all considered values of $\nu$ coloured by 1) $\nu$ value and 2) $k$-means clustering of the minima in high-dimensional space. No clusters are recognisable to distinguish between different values of $\nu$.}
   \label{fig:pcaresults}
\end{figure}

We can show that differences between minima do not exist at the level of individual values of $\nu$. Different values of $\nu$ cannot be distinguished when minima from various $\nu$ runs are shown in a single plot (Fig. \ref{fig:pcaresults}).

This result should not be a surprise since the Mat\'ern function is smooth and hence one expects smooth transitions in -lml values for adjacent, subsequent parameterisations of the kernel. Furthermore, the right plot of Figure \ref{fig:pcaresults} underlines the fact that a 2-dimensional representation is insufficient to characterise the hyperparameter space which is also expected, since every hyperparameter in a GP should be relevant, in contrast to an overparameterised neural network.

\subsection{Computing occupation probability and Hessian spectral norm}\label{app_occprob}
The occupation probability of a minimum is computed by the harmonic approximation to the vibrational density of states, or entropy in molecular sciences. This is a function of the log product of positive Hessian Eigenvalues, given for minimum $\alpha$ by: 
\begin{equation}
p_\alpha(T)=\frac{n_\alpha \exp(-L_\alpha / k_B T) / \bar{\nu}_\alpha^\kappa}{\sum_\gamma n_\gamma \exp(-L_\gamma / k_B T) / \bar{\nu}_\gamma^\kappa}
\end{equation}
where $n_\alpha$ is the log product of positive Hessian eigenvalues $\nu_\alpha$ the geometric mean normal mode vibrational frequency, $L_\alpha$ the loss value (potential energy in molecular sciences for minimum $\alpha$, $T$ the temperature (fictitious in ML systems), $k_B$ is the Boltzmann constant, and $\kappa = 3N-6$ is the number of vibrational degrees of freedom.
Simplified, in ML terms, this becomes: 
\begin{equation}
    p_\alpha(T)=\frac{n_\alpha \exp(-L_\alpha)}{\sum_\gamma n_\gamma \exp(-L_\gamma)},
\end{equation}
only a function of the log product of positive Hessian eigenvalues and the loss.
\subsubsection{Spectral norm}
The spectral norm of a matrix $A$, shown as $||A||$, is the maximum singular value of $A$. Usually, this is the square root of the maximum eigenvalue. However, in the case of a symmetric matrix like the Hessian $H$, the spectral norm $||H||$ is simply the maximum eigenvalue.

\subsection{Normalising weights for GP ensembles}
\begin{figure}[!h]
\centering
  \begin{subfigure}[t]{0.03\linewidth}
    (a)
  \end{subfigure}
  \begin{subfigure}[t]{0.45\linewidth}
    \includegraphics[width=\linewidth, valign=t]{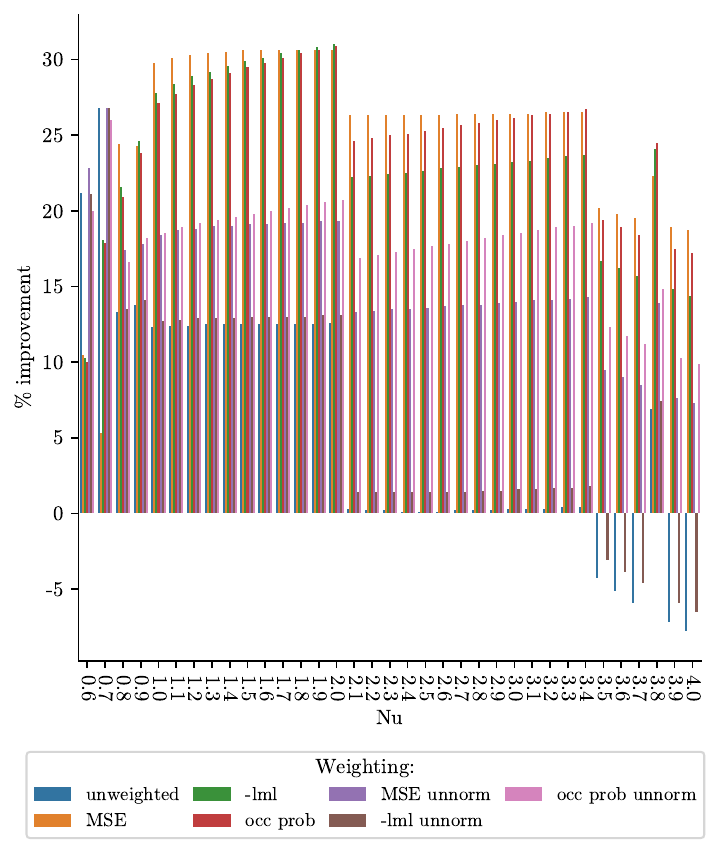}
  \end{subfigure}\hfill
  \begin{subfigure}[t]{0.03\linewidth}
    (b)
  \end{subfigure}
  \begin{subfigure}[t]{0.45\linewidth}
    \includegraphics[width=\linewidth, valign=t]{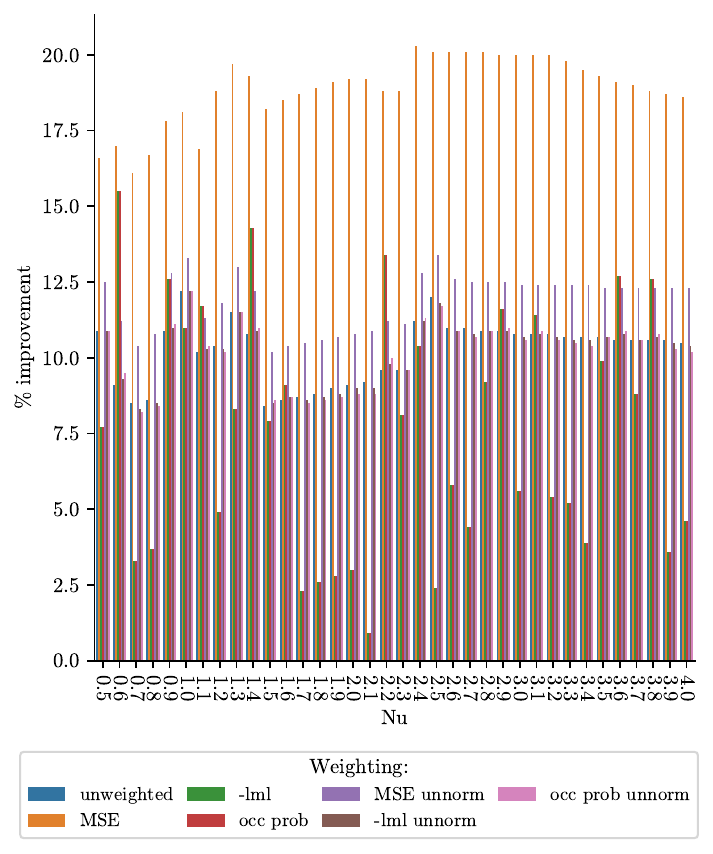}
  \end{subfigure}
   \caption{Percentage improvement over single best minimum for GP ensembles for 3$d$ and 4$d$ Schwefel data. Different weighting strategies are highlighted in the respective colours.}
   \label{fig:ensemble_comparison_withoutnorm}
\end{figure}

Normalisation of weights for GP ensembles is required to improve quality of the ensembles. In Figure \ref{fig:ensemble_comparison_withoutnorm} we compare normalised and non-normalised weighting schemes. Normalisation substantially improves the ensembles.

\begin{SCfigure}[0.5][!h]
\centering
\caption{Difference between normalised and non-normalised weighting of GP ensembles. Let the improvement of normalised ensembles for some weighting scheme be $\delta\mathcal{L}_n^w$ and the improvement of unweighted ensembles be $\delta\mathcal{L}_u^w$, then this figure is computed as $\delta\mathcal{L}_n^w - \delta\mathcal{L}_i^w \forall w$}
\includegraphics[width=.6\linewidth]{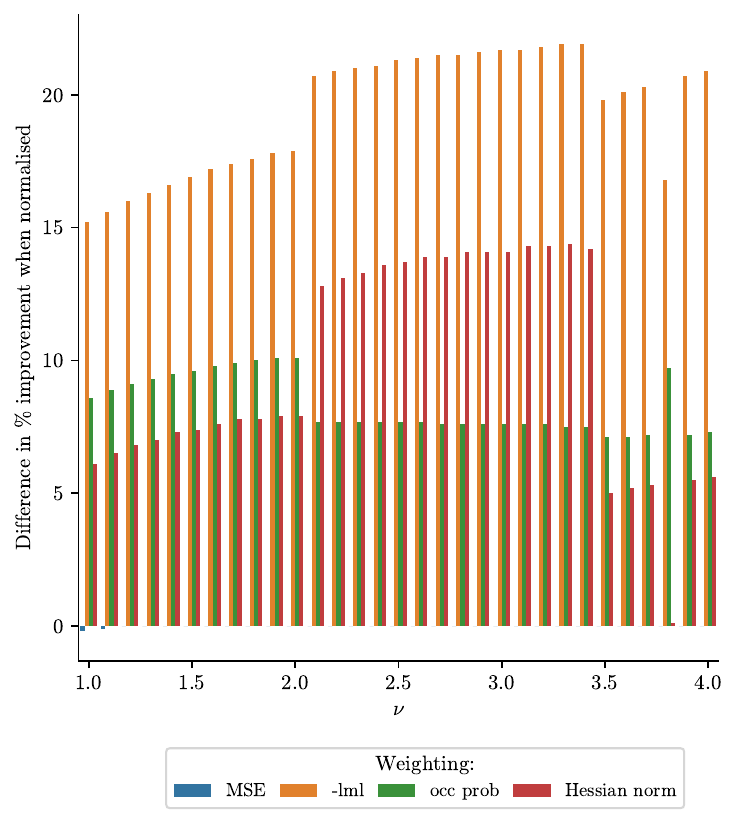}
    \label{fig:normdiff}
\end{SCfigure}

In Figure \ref{fig:normdiff} we make this result even more explicit by showing the absolute difference in \% improvement between normalised and unnormalised weighting. For all values of $\nu$, the improvement by normalisation is striking. 

\begin{figure}[!h]
\centering
  \begin{subfigure}[t]{0.03\linewidth}
    (a)
  \end{subfigure}
  \begin{subfigure}[t]{0.45\linewidth}
    \includegraphics[width=\linewidth, valign=t]{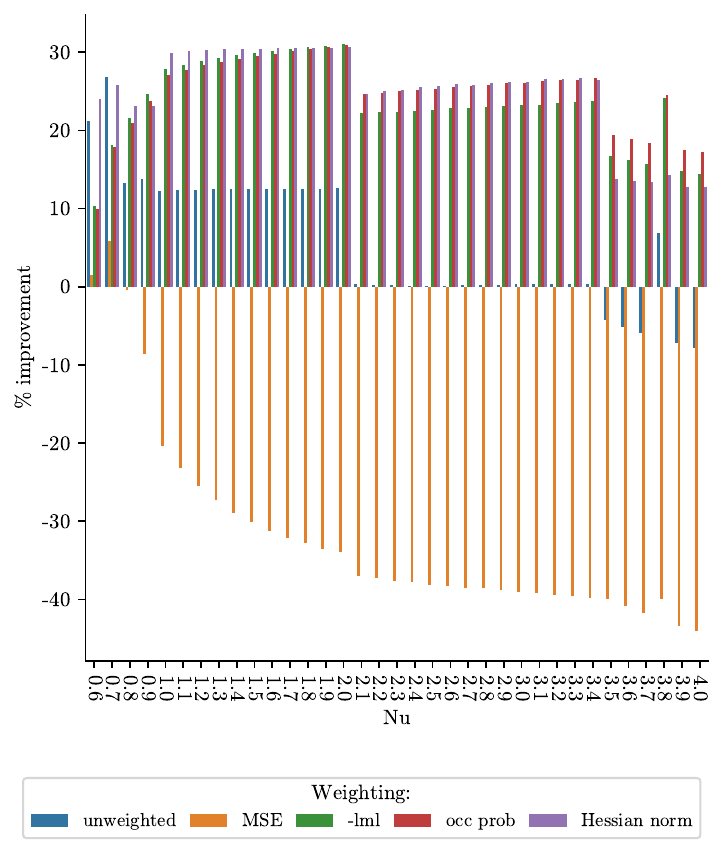}
  \end{subfigure}\hfill
  \begin{subfigure}[t]{0.03\linewidth}
    (b)
  \end{subfigure}
  \begin{subfigure}[t]{0.45\linewidth}
    \includegraphics[width=\linewidth, valign=t]{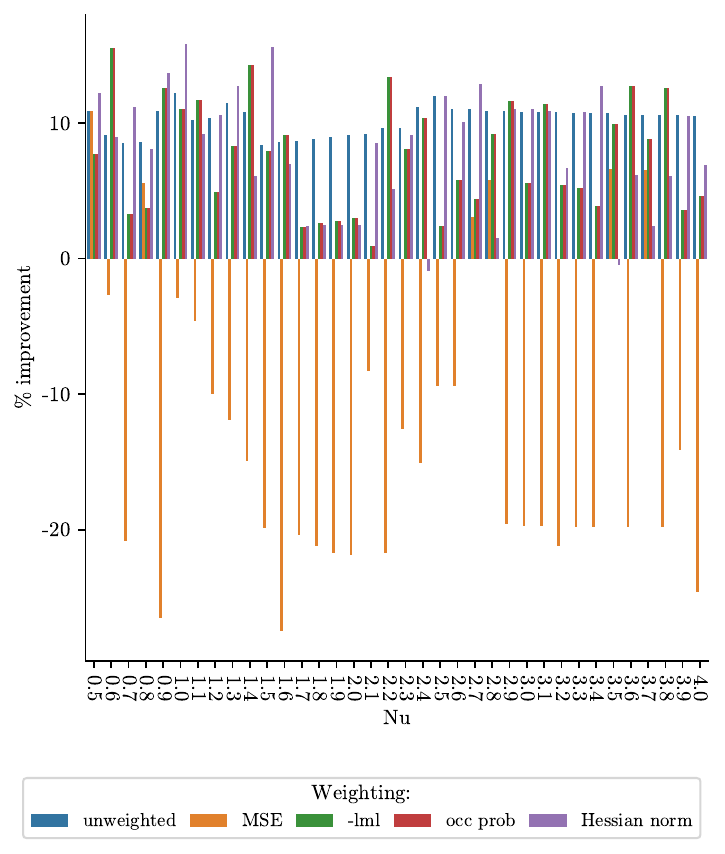}
  \end{subfigure}
   \caption{Percentage improvement over single best minimum for GP ensembles for 3$d$ and 4$d$ Schwefel data. Different weighting strategies are highlighted in the respective colours.}
   \label{fig:train_mse_included}
\end{figure}

Simply weighting the minima by (inverse) training MSE does not yield good results, as can be seen in Figure \ref{fig:train_mse_included}. Weighting by inverse training MSE in fact substantially decreases performance compared to the single best minimum. The explanation is that for any minima considered below, the training MSE lies between $10^{-8}-10^{-13}$. Hence, scaling predictions by these weights means essentially setting the weight for one minimum to be unity, and all others to be 0. 

Thus, weighting by training MSE achieves the opposite to the desired effect. Only a single minimum is considered and since the best minimum by training MSE is not necessarily the one generalising the best, this discrepancy arises. This point is underlined by figure \ref{fig:normdiff} which shows that normalising the weights has no influence on the, already poor, performance of weighting by testing MSE. The values are all small, yet differ in orders of magnitude, which makes weighting with a systematic scheme nearly impossible. 

\end{document}